%% file: 0285.tex
\documentclass[runningheads]{llncs}
\usepackage{graphicx}
\usepackage{amsmath,amssymb} 
\usepackage{expl3}
\usepackage{multirow}

\newcommand{\argmin}{\mathop{\arg\!\min}}

\ExplSyntaxOn
\newcommand\latinabbrev[1]{
  \peek_meaning:NTF . {
    #1\@}%
  { \peek_catcode:NTF a {
      #1.\@ }%
    {#1.\@}}}
\ExplSyntaxOff

\newcommand\blfootnote[1]{%
  \begingroup
  \renewcommand\thefootnote{}\footnote{#1}%
  \addtocounter{footnote}{-1}%
  \endgroup
}

\def\ie{\latinabbrev{\emph{i.e}}}
\def\eg{\latinabbrev{\emph{e.g}}}
\def\etal{\latinabbrev{\emph{et al}}}
\def\etc{\latinabbrev{\emph{etc}}}

\begin{document}

\title{Evolvement Constrained Adversarial Learning for Video Style Transfer} 
\titlerunning{Evolvement Constrained Adversarial Learning} 


\author{First Author\inst{1}\orcidID{0000-1111-2222-3333} \and
Second Author\inst{2,3}\orcidID{1111-2222-3333-4444} \and
Third Author\inst{3}\orcidID{2222--3333-4444-5555}}
\author{
  $\textrm{Wenbo Li}^{1*} \quad \textrm{Longyin Wen}^{2*} \quad \textrm{Xiao Bian}^{3} \quad \textrm{Siwei Lyu}^{1}$
}

%

\authorrunning{Li, Wenbo et al.} 


\institute{$\textrm{}^{1}$University at Albany, SUNY$\quad\textrm{}^{2}$JD Finance AI Lab$\quad\textrm{}^{3}$GE Global Research\\
\email{$\{$wli20,slyu$\}$@albany.edu, lywen.cv.workbox@gmail.com, xiao.bian@ge.com}}

\maketitle

\blfootnote{* indicates equal contributions.}
\input{1_abstract.tex}
\input{2_intro.tex}
\input{3_related.tex}

\input{4_overview.tex}
\input{5_loss.tex}
\input{6_model.tex}
\input{7_experiments.tex}
\input{8_conclusion.tex}

\bibliographystyle{splncs04}
\bibliography{reference}
\end{document}

%% file: 1_abstract.tex
\vspace{-3mm}
\begin{abstract}
   Video style transfer is a useful component for applications such as augmented reality, non-photorealistic rendering, and interactive games. Many existing methods use optical flow to preserve the temporal smoothness of the synthesized video. However, the estimation of optical flow is sensitive to occlusions and rapid motions. Thus, in this work, we introduce a novel evolve-sync loss computed by evolvements to replace optical flow. Using this evolve-sync loss, we build an adversarial learning framework, termed as Video Style Transfer Generative Adversarial Network (VST-GAN), which improves upon the MGAN method for image style transfer for more efficient video style transfer. We perform extensive experimental evaluations of our method and show quantitative and qualitative improvements over the state-of-the-art methods.
\end{abstract}

%% file: 2_intro.tex
\section{Introduction}

\begin{figure}
  \centering
  \includegraphics[width=0.87\linewidth]{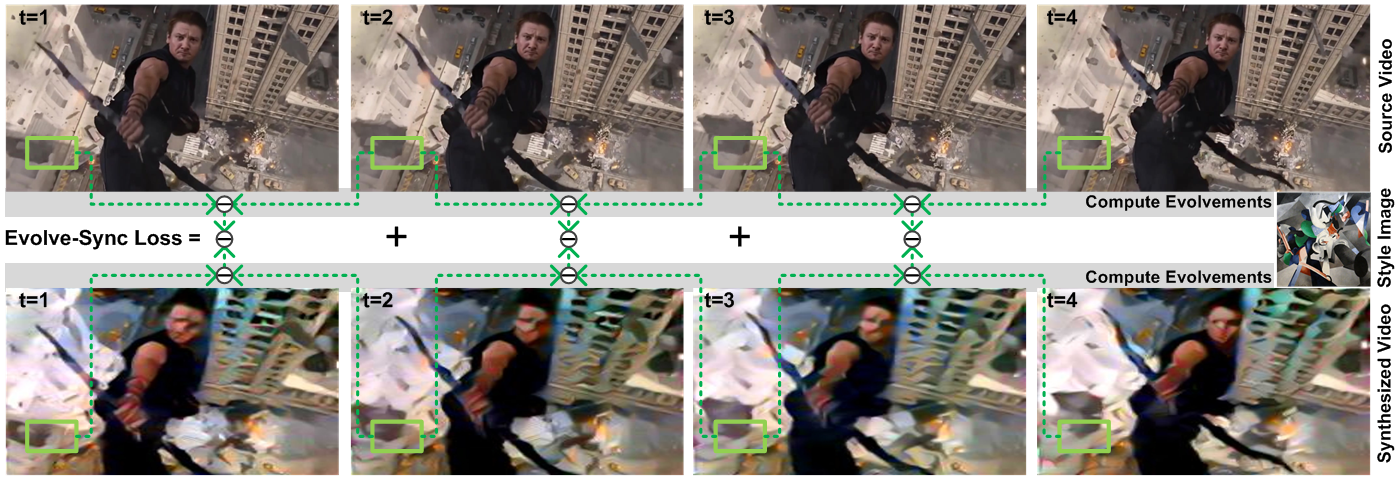}\\ 
  \caption{\em \small
Given a video and an image, our algorithm aims to synthesize a video combining the style of the image and the content of the video. To preserve the temporal smoothness of the synthesized video, we use {\em evolvements} derived from the source and synthesized video, and further compute the evolve-sync loss as the replacement of the optical flow constraints. This loss ensures that the textures at the same location in the image plane of the source and synthesized video evolve synchronously. For the illustration purpose, we only show the order two loss of the evolvements at one patch.
}
\label{Fig1}
\end{figure}

Great artists in history can render scenes with their distinct styles. It is the unique artistic style that differs Van Gogh from Picasso. We wonder if an algorithm can also acquire such styles? For instance, would it be able to re-render the scenes in The Avengers (2012) as if it were the oeuvre of Francis Picabia? Such an interesting question can be formulated as the video style transfer problem as shown in Fig.~\ref{Fig1}, \ie, given a \emph{style image} (Francis Picabia's Udnie) and a \emph{source video} (a clip from The Avengers), the ``synthesizer'' should automatically produce a video combining both the style of Udnie and the content of The Avengers. Such an algorithm can find applications in many areas, such as augmented reality, computer games and nonphotorealistic rendering.

Many recent works in the computer vision and computer graphics community have focused on the problem of image style transfer \cite{GatysEB15,GatysEB16,JohnsonAF16,LiW16,UlyanovLVL16}. However, these methods cannot be readily extended to videos, since independently generating each video frame leads to artifacts such as flickering and jagging in the synthesized videos. To this end, existing video style transfer algorithms \cite{AndersonBMO16,ChenLYYH17,RuderDB16} rely on signals that are estimated by a given motion model such as optical flows computed from adjacent frames to preserve temporal smoothness. We call such signals as model-driven signals. Although more visually pleasing results are achieved with these methods, optical flow estimation methods are known to be sensitive to occlusions and rapid and abrupt motions \cite{RevaudWHS15,WeinzaepfelRHS13}, and such limitations affects the qualities of the synthesized videos. Two recent methods \cite{ChenLYYH17,RuderDB16} attempt to remedy these problems by introducing occlusion masks to filter out low-confidence optical flow, but the generation of occlusion masks is also error-prone and can lead to further artifacts.

In this work, we aim to exploit model-free signals (against the model-driven ones) in the source video for video style transfer, and synthesize video to match with such signals in the source video to preserve the temporal smoothness. To this end, we introduce \textbf{evolvements}, a form of inter-frame variations, as such a model-free signal, the acquisition of which is illustrated in Fig.~\ref{Fig1}. As the source and synthesized videos are synchronous in time, it is natural to require that the textures in the source and synthesized videos evolve synchronously, which we term as the \emph{evolve-sync assumption}. The evolve-sync assumption is incorporated in our method with the \emph{evolve-sync loss}, which encourages the evolvements from the source domain and those from the synthesized domain to be the same. As we need to preserve the temporal smoothness at both the microscopic and macroscopic levels, we extend the evolve-sync loss to be multi-level by regarding the evolvements as distributions and employing encoders (\eg, a pre-trained CNN) to extract samples from these probability distributions. Thus, the evolve-sync loss encourages samples of different distributions at the corresponding level to be the same. We use the maximum mean discrepancy (MMD) \cite{GrettonBRSS12} as the distance measure between probability distributions.

The evolve-sync loss can be combined with an image style transfer method to form the basis of video style transfer algorithms. We choose a state-of-the-art image style transfer method, \ie, Markovian Generative Adversarial Network (MGAN) \cite{LiW16},
\begin{figure}
  \centering
  \includegraphics[width=0.65\linewidth]{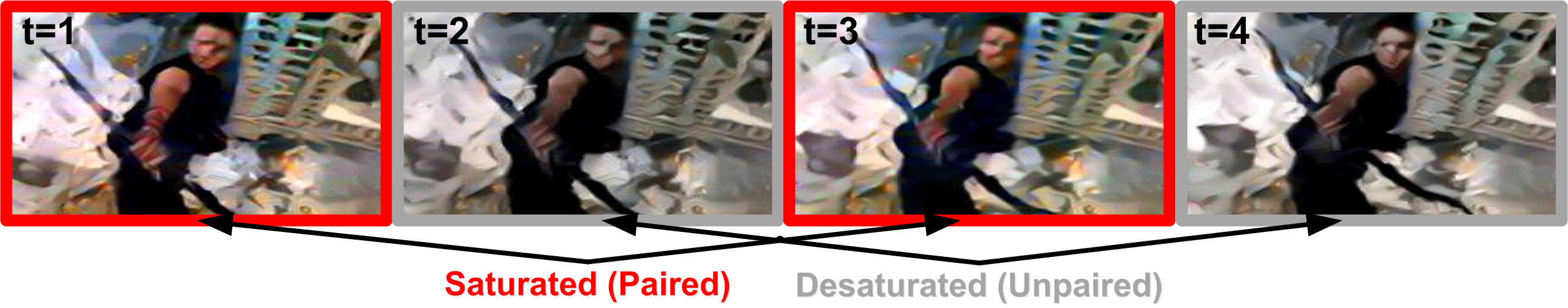}\\ 
  \caption{\em \small
{\bf Saturated vs. Desaturated.}
The desaturated color occurs in the synthesized results for the unpaired frames.
}
\label{Fig2}
\end{figure}
and develop the \emph{Video Style Transfer Generative Adversarial Network} (VST-GAN). MGAN consists of two major components: (i) A Markovian Deconvolutional Adversarial Network (MDAN) denoted by $D$, and (ii) a generator that is a feed-forward convolutional neural network and denoted by $G$. $G$ synthesizes the frames of video according to the content of the source video and the style of the style image, while $D$ plays two roles: it creates real training samples for $G$ with a deconvolutional process driven by the adversarial training, and acts as the adversary to $G$. Besidess the evolve-sync loss, our modifications to MGAN are presented as follows.

As noted in \cite{LiW16}, generating images using the MDAN model $D$ can be slow, and this becomes more problematic for video synthesis. Thus, we design an accelerating training strategy for VST-GAN. Specifically, we only apply $D$ to every other frame to generate real training samples for $G$, leading to that the real training samples are unpaired with the synthesized frames. We expect that $G$ can synthesize desirable textures for the unpaired frames. However, we observe the desaturated color (similar problem arose in \cite{ZhangIE16}) in the synthesized results of the unpaired frames, which is illustrated in Fig.~\ref{Fig2}. We therefore modify $G$ by adding a convolutional recurrent layer as its final output layer, which alleviates the desaturation problem, as the recurrent connection makes it possible to propagate the saturation of the paired frames to the unpaired ones.

The main contributions of our work can be summarized as follows: we introduce the evolve-sync loss, which is based on the evolvement that is more reliable than the estimated optical flow in preserving the temporal smoothness of the synthesized video. Applying the evolve-sync loss at both the microscopic and microscopic levels, we develop VST-GAN, an adversarial learning framework for video style transfer, by adapting the MGAN image style transfer method. Specifically, we add a convolutional recurrent layer as the output layer to resolve the desaturation problem in the synthesized video, which is caused by the trade-off between the training speed and the sufficiency of the real samples. Experimental results demonstrate the effectiveness of the evolve-sync loss and VST-GAN.

%% file: 3_related.tex

\section{Related Works}
{\flushleft \textbf{Image and Video Style Transfer.}} There has been an extensive literature on image style transfer methods, which synthesize images based on sampling low-level features in the given source and style images. The extensions to video style transfer \cite{BousseauNTS07,HaysE04,LuSF10,ZhangLHM11} rely on optical flow to maintain the temporal smoothness of sampling. See \cite{KyprianidisCWI13} for a comprehensive survey.

Recently, deep neural networks have been proved effective for both image \cite{ChenYLYH17,FanCYHYC18,GatysEB15,GatysEB16,HeCLSY18,JohnsonAF16,LiW16,UlyanovLVL16} and video \cite{AndersonBMO16,ChenLYYH17,RuderDB16} style transfer. Gatys \etal~\cite{GatysEB15,GatysEB16} used the convolutional neural network (CNN) to model the patch statistics with a global Gaussian model of the higher-level feature vectors (\eg, activations of CNN), and transferred the style by minimizing the feature reconstruction loss in an iterative deconvolutional process. Two follow-up works, \ie, Johnson \etal~\cite{JohnsonAF16} and Ulyanov \etal~\cite{UlyanovLVL16}, proposed fast implementations of Gatys \etal's method. Both methods employed precomputed decoders trained with a perceptual style loss and obtained significant runtime benefits. In contrast to these three works, Li and Wand \cite{LiW16} argued that real-world contextually related patches do not always comply with a Gaussian distribution, but a complex nonlinear manifold, and proposed MGAN, where a feed-forward generator is adversarially learned to project the contextually related patches to the manifold of patches.

Anderson \etal~\cite{AndersonBMO16} extended Gatys \etal's method to video style transfer. To preserve the temporal smoothness of the synthesized video, they used optical flow to initialize the style transfer optimization, and incorporated the flow explicitly into the loss function. To further reduce artifacts at the boundaries and occluded regions, Ruder \etal~\cite{RuderDB16} introduced masks to filter out optical flows with low confidences in the loss function. Chen \etal~\cite{ChenLYYH17,ChenYLYH18} extended Johnson \etal's method to a feed-forward network for video style transfer. To preserve the temporal smoothness, this method first obtained the current result via a learned flow, and then reduced the artifacts at the occluded regions by fusing the warped result with the independently synthesized result via a learned occlusion mask. In summary, all existing video style transfer methods rely on using optical to preserve temporal smoothness, and use the occlusion mask to stabilize the results. As such, these methods suffer from the common problems in estimating optical flow, \ie, the sensitivity to occlusion and abrupt motion in video.

{\flushleft \textbf{Generative Adversarial Network (GAN).}} GANs \cite{GoodfellowPMXWOCB14} have achieved impressive results for various tasks in image processing, such as style transfer \cite{LiW16}, generation \cite{DentonCSF15}, editing \cite{ZhuKSE16}, representation learning \cite{MathieuZZRSL16}, and translation \cite{ZhuPIE17}, \etc. The key to GANs' success is the idea of an adversarial loss that forces the generated images/videos to be indistinguishable from the real ones. Only a few works develop GANs for videos, \ie, generation \cite{ZhangXL17,XuZHZGHH18,abs-1710-10916} and prediction \cite{MathieuCL15,VondrickPT16}. The real samples of the existing GANs for video generation and prediction are available. However, this is not the case for video style transfer. This is because the qualified real samples for this task should contain both the desirable style and the required content. In VST-GAN, we generate such samples with the deconvolutional model in MGAN constrained by the evolve-sync loss. However, the iterative deconvolutional optimization for videos is slow. Thus, we design a strategy to accelerate the training process of the GAN framework for video style transfer while maintaining the quality of the synthesized videos.



%% file: 4_overview.tex
\begin{figure*}
  \centering
  \includegraphics[width=1\linewidth]{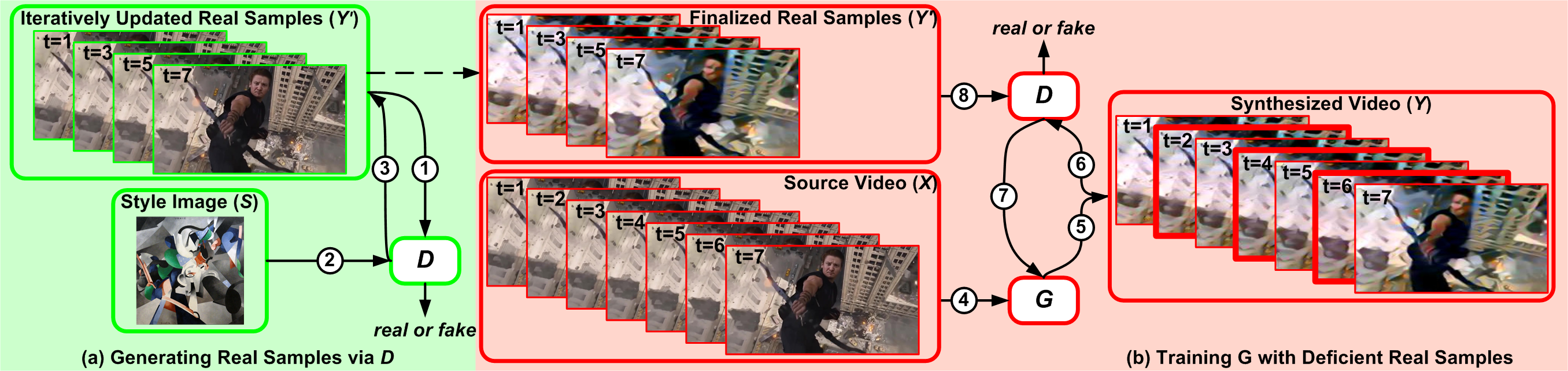}\\ 
  \caption["Short" caption without tikz code]{\em \small
{\bf Pipeline.}
Our method first uses a MDAN model $D$ to generate real samples every other frame within an iterative deconvolutional process. $\mathcal{Y}^{\prime}$ is initialized with the downsampled source video $\mathcal{X}^{\prime}$. The generated real samples are used to train $G$ using a GAN model. Then, $G$ transfers the image style to the whole video. The numbers on the arrows indicate the order of operations, which are explained in detail in $\S$~\ref{sec:overview}.}
\label{Fig3}
\end{figure*}

\section{Overview}
\label{sec:overview}


We first formally define the video style transfer problem as following: given a source video ${\cal X}=\{{\it X}_1, \cdots, {\it X}_i, \cdots \}$ and a style image $S$, we aim to produce a video ${\cal Y}=\{{\it Y}_1, \cdots, {\it Y}_i, \cdots \}$ with the style of $S$, and the content of ${\cal X}$.


We design VST-GAN, an adversarial learning framework based on MGAN \cite{LiW16}, to build a video style transfer algorithm incorporating the evolve-sync loss, with an aim to preserve temporal smoothness of frames without using optical flow. VST-GAN consists of a deconvolutional model $D$, and a feed-forward generator $G$. Both $D$ and $G$ are integrated with the evolve-sync loss. Fig.~\ref{Fig3} illustrates the overall pipeline of our method, which includes two steps:

\noindent \textbf{Step (i)} $D$ generates real samples for $G$ within a deconvolutional process that is constrained by the evolve-sync loss and driven by the adversarial training. Considering the efficiency issue, we accelerate the generation process by applying it to every other frame. In Fig.~\ref{Fig3}~(a), steps \textcircled{1} and \textcircled{2} correspond to the convolutional forward pass, where $D$ determines how real $\mathcal{Y}^{\prime}$ is. Step \textcircled{3} represents the deconvolutional backward pass, where $D$ acts as the generator and the losses are back-propagated to pixels of $\mathcal{Y}^{\prime}$.

\noindent \textbf{Step (ii)} Given the unpaired training samples, $G$ is trained to transfer the style of $S$ to the generated video, using $D$ as the adversary in the manner of GAN. However, the original generator in MGAN suffers from the lack of real samples, which can cause the desaturation effect (or grey image tone) in the generated videos (see an example in Fig.~\ref{Fig2}). We therefore modify $G$ by adding a convolutional recurrent layer as its final output layer, which reduces the desaturation issue effectively. In Fig.~\ref{Fig3}~(b), in the runtime of updating $G$, ${\cal X}$ is fed into $G$ (step \textcircled{4}) to generate ${\cal Y}$ (step \textcircled{5}). Then, $D$ determines how real is the synthesized video ${\cal Y}$ (step \textcircled{6}), and the losses are back-propagated to update $G$ (step \textcircled{7}). During the updating of $D$, ${\cal Y}$ and $\mathcal{Y}^{\prime}$ are used as real and fake samples\footnote{The naming fashion of real and fake samples follow the convention of GAN: the output ${\cal Y}$ of $G$ is considered to be fake, while the precomputed $\mathcal{Y}^{\prime}$ is real.} to train $D$ (steps \textcircled{6} and \textcircled{8}), respectively.


The evolve-sync loss is based on a more reliable signal than the optical flow estimated from the input video, thus it can better preserve the temporal smoothness. Our accelerating training strategy and the added convolutional recurrent structure effectively reduce the training complexity of VST-GAN.

%% file: 5_loss.tex
\section{The Evolve-Sync Loss}
\label{sec:loss}


One basic requirement of video style transfer is to preserve the temporal smoothness between generated frames, as human visual systems are sensitive to the flickering artifacts. This means that the simple approach of generating each frame independently using existing image style transfer algorithms is not effective, as it will lead to visually displeasing results due to two factors. First, as many image style transfer methods (\eg, \cite{GatysEB15,GatysEB16}) are iterative, their results are affected by different initializations and the local minima of the style loss function. Second, a small perturbation in the source images may cause large variations in the synthesized results that are not temporally smooth.

\begin{figure}
  \centering
  \includegraphics[width=0.9\linewidth]{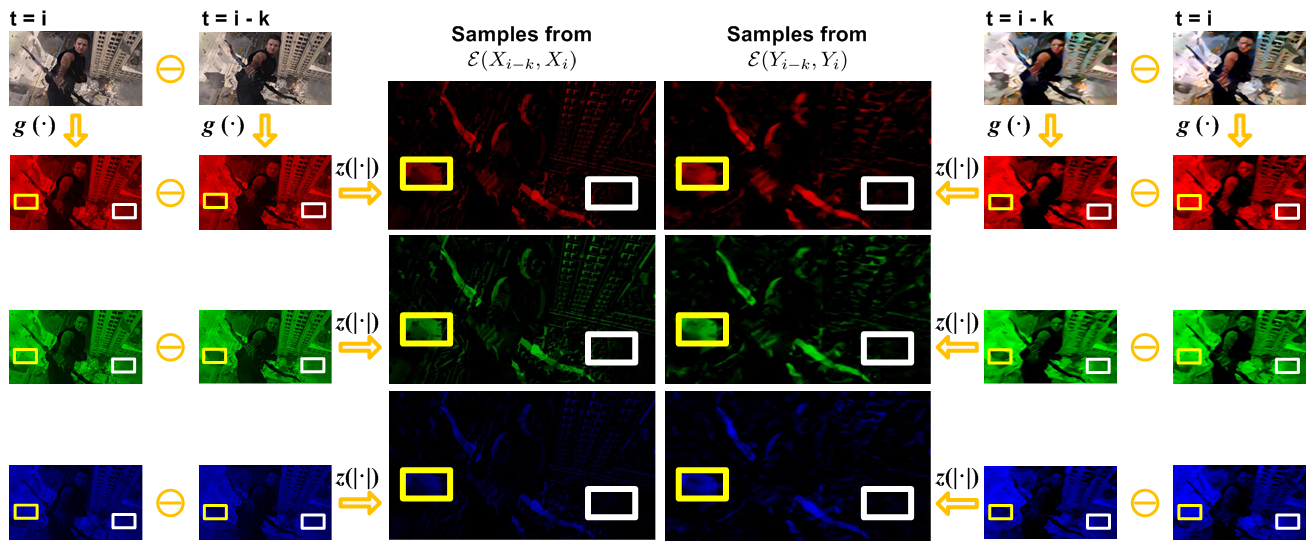}\\ 
  \caption{\em \small
{\bf Illustration of the computation of evolvement and evolve-sync assumption.}
$g(\cdot)$ represents an encoder, which splits the image in R, G and B color channels herein. $z(\cdot)$ represents a standardization function meaning subtracting mean and dividing by standard deviation. $\mathcal{E}(X_{i-k}, X_{i})$ represents the evolvement from frame $X_{i-k}$ to $X_{i}$. The yellow/white boxes highlight a spot in the image plane where drastic/mild variations occur.
}
\label{Fig4}
\end{figure}

As such, in order to generate temporally smooth frames with spatially rich style patterns, existing methods \cite{AndersonBMO16,ChenLYYH17,RuderDB16} modify the image style transfer algorithms by incorporating optical flows estimated from the source video as supervisory signals. The reliability of the estimated optical flow is often problematic due to the problems related with the common optical flow algorithms, \ie, sensitivity to occlusions and rapid motions. This motivates us to turn to a different source of model-free signal directly acquired from the source video itself to capture inter-frame variations, which we term as \emph{evolvement}. Given two frames $X_{i}$ and $X_{i-k} \in\mathbb{R}^{h\times{w}\times{3}}$, we define the evolvement from $X_{i-k}$ to $X_{i}$ as a distribution $\mathcal{E}(X_{i-k}, X_{i})$. Fig.~\ref{Fig4} illustrates the sampling process from evolvements. We compute an evolvement sample $\mathcal{E}(X_{i-k}, X_{i})_{m} \sim \mathcal{E}(X_{i-k}, X_{i})$ as:
\begin{equation}
\label{eq:evolvement}
\footnotesize
\mathcal{E}(X_{i-k}, X_{i})_{m} = z(|g(X_{i})_{m} - g(X_{i-k})_{m}|),
\end{equation}
where $g(\cdot)$ denotes an encoder function that extracts samples from evolvements. The standardization function is represented as $z(x)=\frac{x-\mu}{\sigma}$, where the input $x$ is a 2D matrix, and $\mu$ and $\sigma$ are the mean and standard deviation of elements in $x$, respectively. Index $m$ indicates the $m$th sample generated by $g(\cdot)$.

Our method is based on the \emph{evolve-sync assumption}, which states that $\mathcal{X}$ and the synthesized video $\mathcal{Y}$ are synchronous in time, so their evolvements, $\mathcal{E}(X_{i-k}, X_{i})$ and $\mathcal{E}(Y_{i-k}, Y_{i})$, can be viewed as two synchronized signals. As seen in Fig.~\ref{Fig4}, the brighter a pixel in an evolvement sample is, the more drastic variation occurs at that pixel. The rationality behind the evolve-sync assumption can be understood by contradiction: if it does not hold for a certain pixel, it means that the drastic variation occurs in $\mathcal{E}(X_{i-k}, X_{i})$ while the mild variation occurs in $\mathcal{E}(Y_{i-k}, Y_{i})$, or vice versa. This suggests that the content at that location has not been properly preserved, which contradicts the problem formulation in $\S$~\ref{sec:overview}.

Given $\mathcal{X}$ with a certain temporal smoothness degree, preserving the evolve-sync is equivalent to forcing the temporal smoothness of $\mathcal{Y}$ to be the same as that of $\mathcal{X}$ . Consequently, we introduce the \emph{evolve-sync loss} $L_{es}$ to enforce the evolve-sync assumption in $\mathcal{Y}$ that measures the distance between $\mathcal{E}(X_{i-k}, X_{i})$ and $\mathcal{E}(Y_{i-k}, Y_{i})$. To this end, we employ the Maximum Mean Discrepancy \cite{GrettonBRSS12} as the metric between two probability distributions:
\begin{equation}
\label{eq:evolve-sync-loss1}
\footnotesize
L_{es}(\mathcal{F}, \mathcal{X}, \mathcal{Y}) = \sum_{|i-j| < \delta} \sup_{f \in \mathcal{F}} (
    \mathbf{E}_{x \sim \mathcal{E}(X_{i}, X_{j})} [f(x)] - \mathbf{E}_{y \sim \mathcal{E}(Y_{i}, Y_{j})} [f(y)] ),
\end{equation}
where $\delta$ is a preset parameter determining the order of $L_{es}$,  $\mathcal{F}$ is a Gaussian kernel and we set to $\delta=3$ in our experiments.


We aim to preserve the temporal smoothness of $\mathcal{Y}$ at the microscopic level where the synthesized textures are temporally continuous, and at the macroscopic level where the synthesized textures and the video content are synchronized. To this end, we use two encoders for each level (i) the microscopic encoder $g_{1}(\cdot)$, which splits the image in R, G and B color channels for the microscopic level, and (ii) the macroscopic encoder $g_{2}(\cdot)$, which is a pretrained VGG network (sampled from \emph{Relu3$\_$1}). As such, the overall evolve-sync loss is given as:
\begin{equation}
\label{eq:evolve-sync-loss2}
\footnotesize
L_{es}(\mathcal{G}, \mathcal{F}, \mathcal{X}, \mathcal{Y}) = \sum_{r=1}^{|\mathcal{G}|} \alpha_{r} \cdot \sum_{|i-j| < \delta} \sup_{f \in \mathcal{F}} (
    \mathbf{E}_{x_{r} \sim \mathcal{E}(X_{i}, X_{j})} [f(x)] - \mathbf{E}_{y_{r} \sim \mathcal{E}(Y_{i}, Y_{j})} [f(y)] ),
\end{equation}
where $x_{r}$ and $y_{r}$ are determined by $g_{r}(\cdot)$, and we set $\alpha_{1}=0.005$ and $\alpha_{2}=100$.


%% file: 6_model.tex
\section{Video Style Transfer GAN (VST-GAN)}
\label{sec:model}

In this section, we describe the architecture of VST-GAN in $\S$~\ref{sec:model:architecture} and the training of VST-GAN in $\S$~\ref{sec:model:training}.

\subsection{Architecture}
\label{sec:model:architecture}

\begin{figure}
  \centering
  \includegraphics[width=0.5\linewidth]{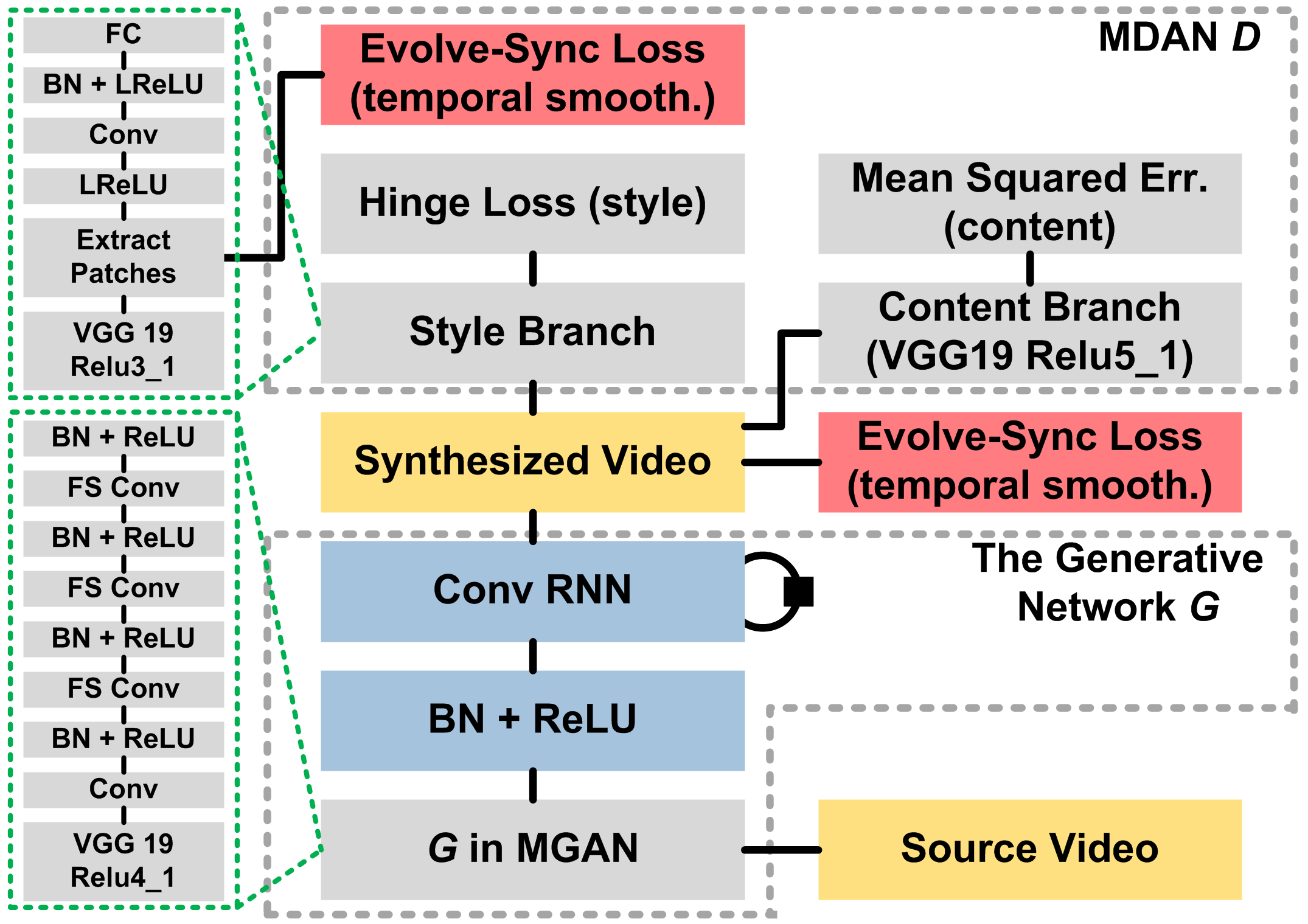}\\ 
  \caption{\em \small
{\bf Architecture of VST-GAN.}
Grey blocks indicate the intrinsic architecture of MGAN, and blocks with other colors indicate the input and our modifications.
}
\label{Fig5}
\end{figure}


We build VST-GAN by adapting Markovian GAN (MGAN) \cite{LiW16}, a state-of-the-art image style transfer framework that does not rely on the implicit assumption that the real-world textures comply with a Gaussian distribution. We show the architecture of VST-GAN in Fig.~\ref{Fig5}, where grey blocks indicate the intrinsic architecture of MGAN, and blocks with other colors indicate our modifications.
MGAN consists of two major components: (i) A Markovian Deconvolutional Adversarial Network (MDAN) denoted by $D$, and (ii) a feed-forward generator denoted by $G$. $D$ plays two roles: it creates real training samples for $G$ with a deconvolutional process that is driven by the adversarial training, and acts as the adversary to $G$.


\noindent \textbf{MDAN $D$.} As shown in Fig.~\ref{Fig5}, $D$ has the \emph{style branch} and the
\emph{content branch}. The style branch learns to distinguish the feature patches extracted from the feature maps output by VGG19 \emph{Relu3$\_$1} of the source video from those of the synthesized videos. $D$ outputs a classification score $s = 1$ or $0$ for each patch, indicating how ``real" the patch is (with $s = 1$ being sampled from the style image $S$, or real patch). For each patch sampled from the synthesized frame, we minimize its style loss (\ie, $1-s$). Like Radford \etal \cite{RadfordMC15}, we use batch normalization (BN) and leaky ReLU (LReLU) to improve the training of $D$. The content branch encourages the content of the synthesized image to be similar to that of the source image, and is constructed from VGG19 features on the same image from higher and more abstract layer \emph{Relu5$\_$1}. The content dissimilarity is measured by a content loss given by the mean squared error between two feature maps obtained from the source video and the synthesized one, respectively. When using $D$ to generate real samples for $G$, the deconvolution process back-propagates both the style and content loss to pixels. When $D$ acts as the adversary to $G$, the style and content loss are back-propagated to train $G$.


\noindent \textbf{The Generative Network $G$.} $D$ requires many iterations and a separate run for each source image, so Li and Wand \cite{LiW16} further developed $G$, which consists of a pre-trained VGG encoder and a decoder. The VGG encoder of $G$ takes the source image as input, and outputs a feature map from \emph{Relu4$\_$1}. The decoder of $G$ takes the output of the encoder, and decodes an image through a ordinary convolution followed by a cascade of fractional-strided convolutions (FS Conv in Fig.~\ref{Fig5}). Note that the content loss is used to measure the content dissimilarity between the synthesized image and its corresponding real sample. Although being trained with fixed-size input, $G$ can be naturally extended to images of arbitrary sizes. VGG encoders in MGAN are fixed during training.


When adapting MGAN to the video style transfer, we make two major modifications to its architecture. First, in order to preserve the temporal smoothness at both the microscopic and macroscopic level, we integrate $D$ with the proposed evolve-sync loss at two levels, \ie, the synthesized video (microscopic) and the VGG encoder of the style branch (macroscopic).

Unfortunately, $D$ has a slow running time -- it takes $D$ nearly 4 hours to synthesize a 50-frame video on a single Titan X GPU -- which is problematic to generate videos with more frames. Thus, we only apply $D$ every other frame to generate real samples for $G$, and train $G$ with such unpaired samples. This way, there will be a half of frames without the corresponding real samples, so these frames will not be used to compute the content loss (inherited from standard regression problems). Since such a content loss encourages conservative predictions, it makes $G$ generate synthesized frames with desaturation artifacts (Fig.~\ref{Fig2}). To alleviate the desaturation problem, we further modify MGAN by adding a convolutional recurrent layer as the final output layer of $G$, as the recurrent connection makes it possible to smooth the saturation of consecutive frames.

\subsection{Training}
\label{sec:model:training}

The training process of VST-GAN includes two steps: (i) generating real training samples for $G$ using $D$ on every other frame, and (ii) training $G$ adversarilly against $D$ with the unpaired training samples.


\noindent \textbf{Generate Real Samples via $D$.} In order to train $G$ adversarially, we need qualified real samples that contain both the style of $S$ and the content of $\mathcal{X}$. As such real samples are not accessible to us during training, we generate them using $D$ on every other frame of the videos as described in $\S$~\ref{sec:overview}. We denote the downsampled source video as ${\cal X^{\prime}}=\{{\it X}_1, {\it X}_3, \cdots, {\it X}_{i-2}, {\it X}_{i}, \cdots \}$ and denote its corresponding real samples as ${\cal Y^{\prime}}=\{{\it Y}_1^{\prime}, {\it Y}_3^{\prime}, \cdots, {\it Y}_{i-2}^{\prime}, {\it Y}_{i}^{\prime}, \cdots \}$. Then, we perform deconvolution with $D$ iteratively to update $\mathcal{Y^{\prime}}$ (initialized with ${\cal X^{\prime}}$), so that the following loss is minimized:
\begin{equation}
\label{eq:D_objective}
\footnotesize
\begin{split}
\hat{\mathcal{Y^{\prime}}} = \argmin_{\mathcal{Y^{\prime}}} &\sum_{Y_{i}^{\prime} \in \mathcal{Y^{\prime}}} [ L_{t}( \Phi_{t}(Y_{i}^{\prime}), \ell_{real} ) + L_{c}( \Phi_{c}(X_{i}), \Phi_{c}(Y_{i}^{\prime})) +\omega \Upsilon (Y_{i}^{\prime}) ] +  \\
&L_{es}(\mathcal{G}, \mathcal{F}, \mathcal{X^{\prime}}, \mathcal{Y^{\prime}}),
\end{split}
\end{equation}
where $L_{t}$ denotes the style loss. $L_{c}$ denotes the content loss, which is a mean squared error. $\Phi_{t}$ and $\Phi_{c}$ denote the VGG encoder in the style and content branch, respectively. $L_{es}$ denotes the evolve-sync loss defined in \eqref{eq:evolve-sync-loss2}. The regularizer $\Upsilon$ is a smoothness prior for pixels \cite{MahendranV15}. We sample patches from $\Phi_{t}(Y_{i}^{\prime})$, and compute $L_{t}$ as the hinge loss with their labels fixed to one, \ie, $\ell_{real}=1$:
\begin{equation}
\label{eq:texture_loss}
\footnotesize
L_{t}( \Phi_{t}(Y_{i}^{\prime}), \ell_{real} ) =  \frac{1}{N} \sum_{j=1}^{N} \max (0, 1 - \ell_{real} \cdot s_{j}),
\end{equation}
where $s_{j}$ denotes the score (output by $D$) of the $j$th patch, and $N$ is the total number of sampled patches in $\Phi_{t}(Y_{i}^{\prime})$.

The model $D$ is trained in tandem: its parameters are randomly initialized, and then updated after each deconvolution, so it improves as $\mathcal{Y^{\prime}}$ improves. The objective of updating $D$ is:
\begin{equation}
\label{eq:D_objective2}
\footnotesize
\hat{D} = \argmin_{D} L_{t}( \Phi_{t}(S), \ell_{real}) + \sum_{Y_{i}^{\prime} \in \mathcal{Y^{\prime}}} L_{t}( \Phi_{t}(Y_{i}^{\prime}), \ell_{fake}).
\end{equation}
$\ell_{real}=1$ and $\ell_{fake}=0$. Like \cite{LiW16}, we set $\omega=0.00001$ in \eqref{eq:D_objective}, and minimize \eqref{eq:D_objective} and \eqref{eq:D_objective2} using back-propagation with ADAM \cite{KingmaB14} (learning rate 0.02, momentum 0.5).
The optimization in \eqref{eq:D_objective2} is memory intensive. To make it feasible and efficient for a machine with a Titan X GPU with 12GB onboard memory, we divide $X^{\prime}$ into multiple non-overlapped segments of 3 frames, and synthesize frames within one segment after another. In this way, $L_{es}$ in \eqref{eq:D_objective} will only preserve the temporal smoothness within each segment. In order to preserve the inter-segment smoothness, we use the last 2 frames of the previous segment to compute $L_{es}$, and leave these 2 frames unchanged during the optimization for the current segment. The segment size can be adaptively enlarged with increased GPU memory capacity.


\noindent \textbf{Train $G$ Against $D$ with Unpaired Real Samples.} Given the unpaired real samples ${\cal Y^{\prime}}$, we aim to train $G$ against $D$ in a GAN model. $G$ takes ${\cal X}$ as input and outputs the synthesized video ${\cal Y}=\{{\it Y}_1, \cdots, {\it Y}_i, \cdots \}$, with $Y_{i} = G(X_{i})$. Thus, our objective herein is as follows:
\begin{equation}
\label{eq:G_objective}
\footnotesize
\begin{split}
 L(G, D, \mathcal{X}, \mathcal{Y}, \mathcal{Y^{\prime}}) = &\sum_{Y_{i} \in \mathcal{Y}} [ L_{t}( \Phi_{t}(Y_{i}), \ell_{real} ) +  \omega \Upsilon (Y_{i}) ] + L_{es}(\mathcal{G}, \mathcal{F}, \mathcal{X}, \mathcal{Y}) + \\
 &\sum_{Y_{i}^{\prime} \in \mathcal{Y^{\prime}}} [ L_{t}( \Phi_{t}(Y_{i}^{\prime}), \ell_{real} ) + L_{c}( \Phi_{c}(Y_{i}), \Phi_{c}(Y_{i}^{\prime})) ].
\end{split}
\end{equation}
We therefore aim to solve:
\begin{equation}
\label{eq:G_objective_solver}
\footnotesize
\hat{G} = \arg \min_{G} \max_{D} L(G, D, \mathcal{X}, \mathcal{Y}, \mathcal{Y^{\prime}}),
\end{equation}
where $D$ and $G$ are trained from scratch using back-propagation with ADAM (learning rate 0.02, momentum 0.5). Same notations as those in \eqref{eq:G_objective} can be found in \eqref{eq:D_objective}, \eqref{eq:texture_loss} and \eqref{eq:D_objective2}. Note that $L_{c}$ is only valid for the paired frames.

%% file: 7_experiments.tex
\section{Experiments}

\begin{figure*}
  \centering
  \includegraphics[width=1\linewidth]{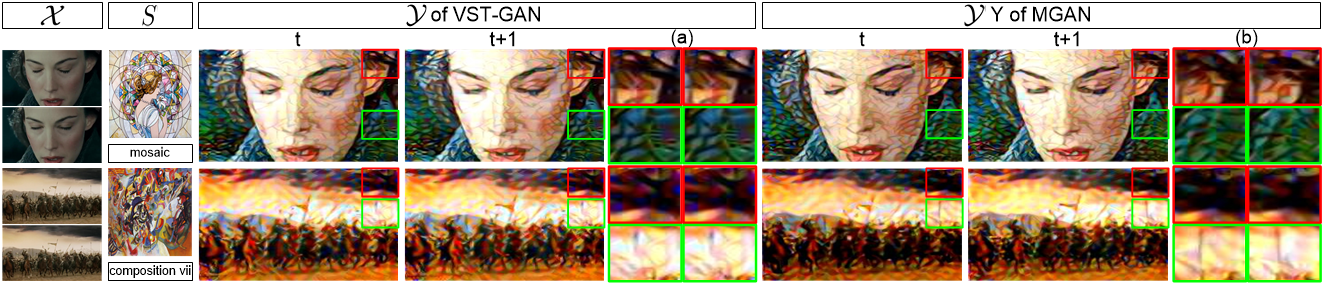}\\ 
  \caption{\em \small
{\bf Qualitative comparison with MGAN \cite{LiW16}} The marked regions highlight that the temporal smoothness of our results is higher. The dilated marked regions are shown in (a) and (b). This figure is best viewed in color. }
\label{Fig6}
\end{figure*}

\noindent \textbf{Implementation Details.} We implement VST-GAN and MGAN using Tensorflow, and conduct the experiments on a computer with an Intel Xeon X5570 CPU with $16$ cores of 2.93GHz each, 94.4GB memory, and one NVIDIA TITAN X GPU with 12GB onboard memory. For the real sample generation process, $D$ is trained for each segment (3 frames) for $3,000$ iterations. With a batch size of 3, $G$ is trained for $20,000$ iterations. For a 50-frame video, it takes $D$ approximately 2 hours to generate the real samples, and approximately a further 1 hour to train $G$.

\noindent \textbf{Datasets.} We use 8 classical style images, \ie, \emph{starry night}, \emph{the scream}, \emph{udnie}, \emph{la muse}, \emph{wave}, \emph{composition vii}, \emph{mosaic}, and \emph{candy}, several of which are used in \cite{ChenLYYH17} or \cite{RuderDB16}. For the source videos, we choose $8$ videos with diverse contents, including natural scenes, action scenes, close-up portraits, \etc. Lengths of these videos vary from 40 to 300 frames, with 91 frames on average. All videos have the image resolutions of 640$\times$360 and were captured at $23$ frame per second.

\noindent \textbf{Compared Methods.} We compare VST-GAN with ASTV \cite{RuderDB16}, a state-of-the-art neural network based video style transfer method. ASTV uses optical flow and occlusion mask to preserve temporal smoothness in the synthesized video, so it suffers from the common problems in estimating optical flow, \ie, the sensitivity to occlusion and abrupt motion in video. We also create a baseline method based on MGAN \cite{LiW16}, which uses image style transfer method in \cite{LiW16} to create individual frames independently. Comparison with these baseline methods demonstrates the advantage of our method in preserving temporal smoothness.

\subsection{Qualitative Comparison}
In Fig.~\ref{Fig6}, we show two consecutive frames from two synthesized videos produced by VST-GAN and MGAN, respectively, with two highlighted regions in each frame. The close-up regions demonstrate the effectiveness of evolve-sync loss in preserving the temporal smoothness. As mentioned in the beginning of $\S$~\ref{sec:loss}, the image style transfer methods (\eg, MGAN) are ineffective in preserving the temporal smoothness, which is evident from comparing the two close-up regions.

\begin{figure*}
  \centering
  \includegraphics[width=1\linewidth]{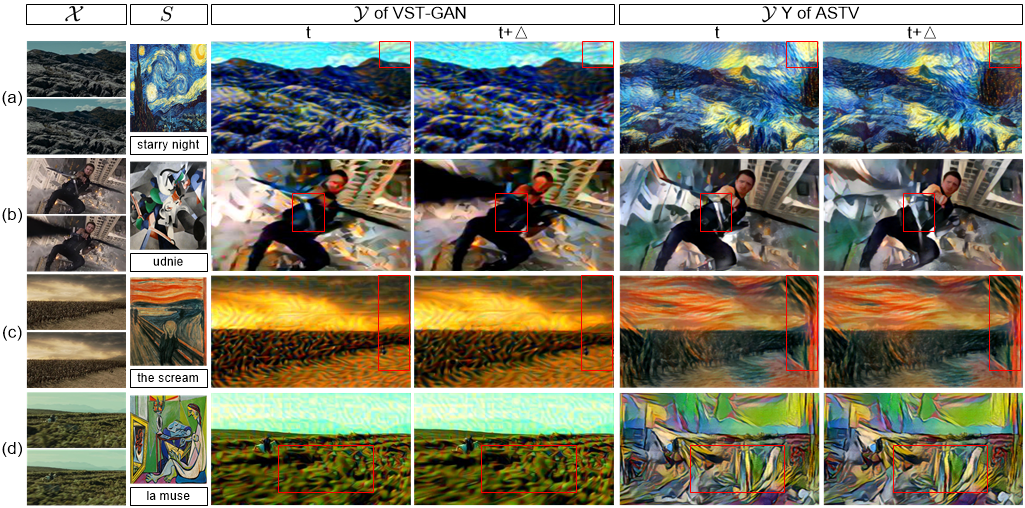}\\ 
  \caption{\em \small
{\bf Qualitative comparison of the results of VST-GAN with those of ASTV \cite{RuderDB16}} Row (a) corresponds to a scene with the rapid camera motion, and we highlight the newly entering regions where the artifacts appear for ASTV's result. Row (b) displays a video associated with the rapid object motions and occlusions. The marked regions show that the artifacts appear after rapid motions and occlusions for ASTV's result. Row (c) presents an example where the ghosting salient content exists in ASTV's result but not in $\mathcal{X}$. Row (d) illustrates an example where the content of $\mathcal{X}$ is not preserved properly by ASTV. This figure is best viewed in color. }
\label{Fig7}
\end{figure*}

We further compare the synthesis results using VST-GAN with ASTV \cite{RuderDB16} in Fig.~\ref{Fig7}, with four video clips with a variety of challenging factors including camera motions, rapid object motions, and occlusions, \etc. Fig.~\ref{Fig7}~(a) shows an epic natural scene with lateral camera motions. Note that artifacts emerge as the new content enters the scene at the top right corner. Considering the contextual information on the left side, the sky color within the marked region is supposed to be either blue, white or yellow. However, the actual color is brownish-grey, which is partially caused by optical flow's intrinsic limitation. Specifically, since the camera view moves from left to right, the estimated direction of the optical flow is the opposite. This leads to an ill-posed problem that the new content at the top right corner solely depends on the pixel values along image's right border. Thus, colors for the sky region (blue, white and yellow) on the left side have no effect on the synthesized video. Fig.~\ref{Fig7}~(b) displays a video in which the arrow moves rapidly and its movement causes the occlusion. Artifacts arise in the videos synthesized by ASTV due to the instability of the estimated optical flow in the presence of rapid motions and occlusions. On the other hand, because we introduce the evolve-sync loss to replace optical flow, VST-GAN is not affected by problems of the optical flow estimation.

In Fig.~\ref{Fig7}~(c), we present an example where ASTV synthesized video contains some ``ghost'' salient content that does not exist in the source video. The preservation of the source video content is worse in the ASTV's result in Fig.~\ref{Fig7}~(d), \eg, the contextually related patches that constitute the grassland in $\mathcal{X}$ become unrelated in $\mathcal{Y}$. This is because ASTV \cite{RuderDB16} models textures in the style image with a Gaussian distribution. As a result, the synthesized video does not further improve once two distribution matches and the synthesis quality of local image regions cannot be guaranteed. In contrast, VST-GAN preserves the content more properly by relaxing the above assumption to that the textures follow a complicated non-linear manifold. Furthermore, the adversarial training of VST-GAN can recognize such a manifold with its discriminative network, and strengthen its generative power with a projection on the manifold.

\subsection{Quantitative Comparison}
\noindent \textbf{Evaluation Metrics.} Existing methods \cite{ChenLYYH17,RuderDB16} measure the temporal smoothness of the synthesized video using the ground-truth optical flow and occlusion mask. Specifically, they warp the $i$-th frame in the synthesized video to be synchronized with the ground truth flow and compute the difference with the $(i-1)$-th synthesized frame in non-occluded regions. Although this metric is straightforward to compute, it has two drawbacks. First, it restricts the choice of the evaluated
\begin{table}
\scriptsize
\centering
\begin{tabular}[t]{ |p{6em}|p{9em}|p{4em}|p{4em}|p{4em}|p{4em}|p{4em}|p{4em}|}\hline
    \multirow{2}{*}{\makebox[6em]{Sequence}} & \multirow{2}{*}{\makebox[9em]{Method}} & \multicolumn{6}{c|}{\makebox[16em]{AESL} } \\ \cline{3-8}
    & & \makebox[4em]{2-order} &\makebox[4em]{4-order} &\makebox[4em]{6-order} &\makebox[4em]{8-order} &\makebox[4em]{10-order} &\makebox[4em]{12-order}\\ \hline
    \multirow{5}{*}{\makebox[6em]{starry~night}} & ASTV \cite{RuderDB16} & \makebox[4em]{\textbf{45.22}} & \makebox[4em]{\textbf{132.74}} & \makebox[4em]{221.95} & \makebox[4em]{301.32} & \makebox[4em]{366.93} & \makebox[4em]{463.47} \\ \cline{2-8}
        & MGAN \cite{LiW16} & \makebox[4em]{97.67} & \makebox[4em]{245.43} & \makebox[4em]{344.66} & \makebox[4em]{443.56} & \makebox[4em]{523.67} & \makebox[4em]{598.88} \\ \cline{2-8}
        & VST-GAN (ours) & \makebox[4em]{60.42} & \makebox[4em]{143.62} & \makebox[4em]{\textbf{220.32}} & \makebox[4em]{\textbf{289.87}} & \makebox[4em]{\textbf{344.98}} & \makebox[4em]{\textbf{412.83}} \\ \cline{2-8}
        & VST-GAN w/o ESL & \makebox[4em]{72.13} & \makebox[4em]{181.80} & \makebox[4em]{274.85} & \makebox[4em]{379.43} & \makebox[4em]{440.05} & \makebox[4em]{533.49} \\ \cline{2-8}
        & VST-GAN w/o RNN & \makebox[4em]{65.36} & \makebox[4em]{175.23} & \makebox[4em]{271.84} & \makebox[4em]{347.59} & \makebox[4em]{375.17} & \makebox[4em]{422.74} \\ \hline\hline
    \multirow{5}{*}{\makebox[6em]{the~scream}} & ASTV \cite{RuderDB16} & \makebox[4em]{32.45} & \makebox[4em]{105.31} & \makebox[4em]{178.48} & \makebox[4em]{237.00} & \makebox[4em]{308.06} & \makebox[4em]{368.63} \\ \cline{2-8}
        & MGAN \cite{LiW16} & \makebox[4em]{91.83} & \makebox[4em]{220.06} & \makebox[4em]{292.33} & \makebox[4em]{362.17} & \makebox[4em]{420.37} & \makebox[4em]{478.57} \\ \cline{2-8}
        & VST-GAN (ours) & \makebox[4em]{\textbf{31.33}} & \makebox[4em]{\textbf{96.74}} & \makebox[4em]{\textbf{139.05}} & \makebox[4em]{\textbf{179.07}} & \makebox[4em]{\textbf{204.43}} & \makebox[4em]{\textbf{240.20}} \\ \cline{2-8}
        & VST-GAN w/o ESL & \makebox[4em]{65.65} & \makebox[4em]{143.11} & \makebox[4em]{197.96} & \makebox[4em]{249.81} & \makebox[4em]{293.99} & \makebox[4em]{348.53} \\ \cline{2-8}
        & VST-GAN w/o RNN & \makebox[4em]{47.62} & \makebox[4em]{108.26} & \makebox[4em]{146.45} & \makebox[4em]{193.29} & \makebox[4em]{221.19} & \makebox[4em]{252.69} \\ \hline\hline
    \multirow{5}{*}{\makebox[6em]{udnie}} & ASTV \cite{RuderDB16} & \makebox[4em]{\textbf{48.38}} & \makebox[4em]{84.83} & \makebox[4em]{120.23} & \makebox[4em]{154.45} & \makebox[4em]{187.89} & \makebox[4em]{220.71} \\ \cline{2-8}
        & MGAN \cite{LiW16} & \makebox[4em]{81.14} & \makebox[4em]{121.00} & \makebox[4em]{149.36} & \makebox[4em]{173.43} & \makebox[4em]{226.53} & \makebox[4em]{252.21} \\ \cline{2-8}
        & VST-GAN (ours) & \makebox[4em]{48.69} & \makebox[4em]{\textbf{72.36}} & \makebox[4em]{\textbf{93.31}} & \makebox[4em]{\textbf{112.74}} & \makebox[4em]{\textbf{131.06}} & \makebox[4em]{\textbf{148.46}} \\ \cline{2-8}
        & VST-GAN w/o ESL & \makebox[4em]{81.00} & \makebox[4em]{116.75} & \makebox[4em]{148.03} & \makebox[4em]{177.20} & \makebox[4em]{204.52} & \makebox[4em]{230.32} \\ \cline{2-8}
        & VST-GAN w/o RNN & \makebox[4em]{54.60} & \makebox[4em]{89.00} & \makebox[4em]{119.84} & \makebox[4em]{148.73} & \makebox[4em]{176.09} & \makebox[4em]{202.08} \\ \hline\hline
    \multirow{5}{*}{\makebox[6em]{la~muse}} & ASTV \cite{RuderDB16} & \makebox[4em]{84.18} & \makebox[4em]{266.25} & \makebox[4em]{459.09} & \makebox[4em]{650.63} & \makebox[4em]{838.49} & \makebox[4em]{1021.58} \\ \cline{2-8}
        & MGAN \cite{LiW16} & \makebox[4em]{193.32} & \makebox[4em]{493.94} & \makebox[4em]{737.08} & \makebox[4em]{940.96} & \makebox[4em]{1120.69} & \makebox[4em]{1282.29} \\ \cline{2-8}
        & VST-GAN (ours) & \makebox[4em]{\textbf{79.40}} & \makebox[4em]{\textbf{234.34}} & \makebox[4em]{\textbf{322.61}} & \makebox[4em]{\textbf{396.55}} & \makebox[4em]{\textbf{480.32}} & \makebox[4em]{\textbf{541.78}} \\ \cline{2-8}
        & VST-GAN w/o ESL & \makebox[4em]{131.70} & \makebox[4em]{309.85} & \makebox[4em]{460.34} & \makebox[4em]{595.70} & \makebox[4em]{721.95} & \makebox[4em]{841.39} \\ \cline{2-8}
        & VST-GAN w/o RNN & \makebox[4em]{96.43} & \makebox[4em]{247.61} & \makebox[4em]{325.74} & \makebox[4em]{413.65} & \makebox[4em]{493.38} & \makebox[4em]{568.11} \\ \hline\hline
    \multirow{5}{*}{\makebox[6em]{wave}} & ASTV \cite{RuderDB16} & \makebox[4em]{59.97} & \makebox[4em]{147.53} & \makebox[4em]{233.54} & \makebox[4em]{321.63} & \makebox[4em]{413.23} & \makebox[4em]{506.86} \\ \cline{2-8}
        & MGAN \cite{LiW16} & \makebox[4em]{180.76} & \makebox[4em]{434.75} & \makebox[4em]{636.53} & \makebox[4em]{813.08} & \makebox[4em]{978.61} & \makebox[4em]{1134.02} \\ \cline{2-8}
        & VST-GAN (ours) & \makebox[4em]{\textbf{58.21}} & \makebox[4em]{\textbf{138.57}} & \makebox[4em]{\textbf{193.25}} & \makebox[4em]{\textbf{254.14}} & \makebox[4em]{\textbf{306.07}} & \makebox[4em]{\textbf{361.79}} \\ \cline{2-8}
        & VST-GAN w/o ESL & \makebox[4em]{130.35} & \makebox[4em]{314.50} & \makebox[4em]{466.62} & \makebox[4em]{603.03} & \makebox[4em]{732.35} & \makebox[4em]{855.11} \\ \cline{2-8}
        & VST-GAN w/o RNN & \makebox[4em]{123.84} & \makebox[4em]{294.62} & \makebox[4em]{429.18} & \makebox[4em]{546.39} & \makebox[4em]{656.17} & \makebox[4em]{759.13} \\ \hline\hline
    \multirow{5}{*}{\makebox[6em]{comp.~vii}} & ASTV \cite{RuderDB16} & \makebox[4em]{\textbf{32.92}} & \makebox[4em]{\textbf{99.67}} & \makebox[4em]{172.53} & \makebox[4em]{234.40} & \makebox[4em]{305.62} & \makebox[4em]{345.03} \\ \cline{2-8}
        & MGAN \cite{LiW16} & \makebox[4em]{94.61} & \makebox[4em]{222.30} & \makebox[4em]{304.92} & \makebox[4em]{357.22} & \makebox[4em]{417.04} & \makebox[4em]{456.96} \\ \cline{2-8}
        & VST-GAN (ours) & \makebox[4em]{39.23} & \makebox[4em]{104.42} & \makebox[4em]{\textbf{157.65}} & \makebox[4em]{\textbf{208.61}} & \makebox[4em]{\textbf{254.18}} & \makebox[4em]{\textbf{293.65}} \\ \cline{2-8}
        & VST-GAN w/o ESL & \makebox[4em]{92.03} & \makebox[4em]{245.12} & \makebox[4em]{373.81} & \makebox[4em]{470.22} & \makebox[4em]{578.90} & \makebox[4em]{642.62} \\ \cline{2-8}
        & VST-GAN w/o RNN & \makebox[4em]{65.00} & \makebox[4em]{159.93} & \makebox[4em]{239.70} & \makebox[4em]{303.14} & \makebox[4em]{354.22} & \makebox[4em]{413.45} \\ \hline\hline
    \multirow{5}{*}{\makebox[6em]{mosaic}} & ASTV \cite{RuderDB16} & \makebox[4em]{\textbf{30.44}} & \makebox[4em]{103.48} & \makebox[4em]{187.32} & \makebox[4em]{275.91} & \makebox[4em]{364.91} & \makebox[4em]{451.67} \\ \cline{2-8}
        & MGAN \cite{LiW16} & \makebox[4em]{99.35} & \makebox[4em]{260.26} & \makebox[4em]{400.07} & \makebox[4em]{528.03} & \makebox[4em]{645.29} & \makebox[4em]{751.34} \\ \cline{2-8}
        & VST-GAN (ours) & \makebox[4em]{36.84} & \makebox[4em]{\textbf{98.88}} & \makebox[4em]{\textbf{157.79}} & \makebox[4em]{\textbf{214.84}} & \makebox[4em]{\textbf{253.38}} & \makebox[4em]{\textbf{301.66}} \\ \cline{2-8}
        & VST-GAN w/o ESL & \makebox[4em]{63.13} & \makebox[4em]{179.14} & \makebox[4em]{289.34} & \makebox[4em]{395.64} & \makebox[4em]{497.12} & \makebox[4em]{591.47} \\ \cline{2-8}
        & VST-GAN w/o RNN & \makebox[4em]{53.84} & \makebox[4em]{144.03} & \makebox[4em]{225.46} & \makebox[4em]{301.84} & \makebox[4em]{374.40} & \makebox[4em]{442.21} \\ \hline\hline
    \multirow{5}{*}{\makebox[6em]{candy}} & ASTV \cite{RuderDB16} & \makebox[4em]{29.98} & \makebox[4em]{90.81} & \makebox[4em]{148.01} & \makebox[4em]{200.58} & \makebox[4em]{249.69} & \makebox[4em]{296.37} \\ \cline{2-8}
        & MGAN \cite{LiW16} & \makebox[4em]{33.88} & \makebox[4em]{84.39} & \makebox[4em]{126.82} & \makebox[4em]{164.90} & \makebox[4em]{200.51} & \makebox[4em]{233.95} \\ \cline{2-8}
        & VST-GAN (ours) & \makebox[4em]{\textbf{19.26}} & \makebox[4em]{\textbf{52.34}} & \makebox[4em]{\textbf{87.74}} & \makebox[4em]{\textbf{110.95}} & \makebox[4em]{\textbf{144.60}} & \makebox[4em]{\textbf{170.47}} \\ \cline{2-8}
        & VST-GAN w/o ESL & \makebox[4em]{31.07} & \makebox[4em]{76.31} & \makebox[4em]{114.84} & \makebox[4em]{149.95} & \makebox[4em]{182.66} & \makebox[4em]{213.52} \\ \cline{2-8}
        & VST-GAN w/o RNN & \makebox[4em]{23.67} & \makebox[4em]{67.95} & \makebox[4em]{107.51} & \makebox[4em]{143.22} & \makebox[4em]{175.31} & \makebox[4em]{203.78} \\ \hline
\end{tabular}
\caption{\em \small Comparison on temporal smoothness for synthetic videos using ours and state-of-the-art video style transfer methods.
}
\label{tab1}
\end{table}
videos to those with ground truth optical flow, which are very difficult to generate and scarce in number. Second, it does not allow for the evaluation of long-term temporal smoothness due to the lack of long-term ground truth optical flow. To this end, we use the \emph{averaging evolve-sync loss} (AESL) (averaged by the video length) as a new metric that is free of the optical flow, occlusion mask and the short-term restrictions. We compute the multi-order AESL to evaluate the temporal smoothness for short (order 2 and 4)/medium (order 6 and 8)/long-term (order 10 and 12).

\noindent \textbf{Comparing with the State-of-the-art Methods.} The comparison results are presented in Table~\ref{tab1}. These result show that VST-GAN outperforms MGAN significantly in terms of the temporal smoothness of the synthesized videos. The comparison between VST-GAN and ASTV based on AESL of order 2 and 4 suggests comparable performance of our method using the evolve-sync loss to those based on optical flow and occlusion mask in preserving the short-term temporal smoothness. In addition, the evolve-sync loss is more effective than optical flow in preserving medium/long-term temporal smoothness, which is demonstrated by the comparison based on AESL of order 6, 8, 10 and 12. This is due the lack of long-term optical flows, which cannot be reliably estimated using current methods. In contrast, the high-order evolve-sync loss can be more easily computed and compared.

\noindent \textbf{Effects of the Evolve-sync Loss.} To investigate the impact of the evolve-sync loss on preserving the temporal smoothness, we remove it from the objective of training $G$ \eqref{eq:G_objective}. As a result, we observe significant increase in AESL of all orders, which indicates the retrogression on the preservation of the temporal smoothness. Nonetheless, VST-GAN still preserves the temporal smoothness better than MGAN even without the use of the evolve-sync loss. This is because we maintain the evolve-sync loss in $D$ for generating real training samples, which further demonstrates the effectiveness of the evolve-sync loss.

\noindent \textbf{Effects of the Recurrent Structure.} We remove the convolutional recurrent layer from VST-GAN to study its impact on preserving the temporal smoothness. Consequently, the AESL increases slightly, but the increment is much smaller compared to that after removing the evolve-sync loss. This indicates that the recurrent structure is also useful for preserving the temporal smoothness, but its impact is less prominent than that of the evolve-sync loss.

\noindent \textbf{Runtime Efficiency.} The runtime speed of our VST-GAN in synthesizing videos is $18.18$ fps, which is comparable to the image style transfer method MGAN ($19.33$ fps), and much efficient than the deconvolutional video style transfer method ASTV ($0.03$ fps).

%% file: 8_conclusion.tex
\section{Conclusion}

In this work, we propose VST-GAN as an adversarial learning framework for video style transfer based on the evolve-sync loss. We show that the evolve-sync loss is able to preserve the temporal smoothness effectively without using optical flow. Our accelerating training strategy and the convolutional recurrent structure significantly reduce the training complexity of VST-GAN. Experimental evaluations show that VST-GAN outperforms the state-of-the-art methods based on optical flow in both running time efficiency and visual quality.